\documentclass[10pt,twocolumn,letterpaper]{article}

\usepackage{iccv}
\usepackage{times}
\usepackage{epsfig}
\usepackage{graphicx}
\usepackage{amsmath}
\usepackage{amssymb}
\usepackage{multirow}

\usepackage[pagebackref=true,breaklinks=true,letterpaper=true,colorlinks,bookmarks=false]{hyperref}

\iccvfinalcopy 

\def\iccvPaperID{7837} 

\ificcvfinal\fi

\begin{document}

\title{OmniPose: A Multi-Scale Framework for Multi-Person Pose Estimation}

\author{Bruno Artacho \hspace{2cm} Andreas Savakis \\
Rochester Institute of Technology\\
Rochester, NY\\
{\tt\small bmartacho@mail.rit.edu \hspace{0.5cm} andreas.savakis@rit.edu}}

\maketitle
\ificcvfinal\thispagestyle{empty}\fi

\begin{abstract}
We propose OmniPose, 
a single-pass, end-to-end trainable framework, that achieves state-of-the-art results for multi-person pose estimation. Using a novel waterfall module, the OmniPose architecture leverages multi-scale feature representations that increase the effectiveness of backbone feature extractors, without the need for post-processing. 
OmniPose incorporates contextual information across scales and joint localization with Gaussian heatmap modulation at the multi-scale feature extractor to estimate human pose with state-of-the-art accuracy.
The multi-scale representations, obtained by the improved waterfall module in OmniPose, leverage the efficiency of progressive filtering in the cascade architecture, while maintaining multi-scale fields-of-view comparable to spatial pyramid configurations.
Our results on multiple datasets demonstrate that OmniPose, with an improved HRNet backbone and waterfall module, is a robust and efficient architecture for multi-person pose estimation that achieves state-of-the-art results.
\end{abstract}

\section{Introduction}
Human pose estimation is an important task in computer vision that has generated high interest for  methods on 2D pose estimation \cite{DeepPose}, \cite{HourGlass},
\cite{CPM}, \cite{UniPose},
\cite{HRNet} and 3D \cite{LCR-Net}, \cite{Monocap}, \cite{DensePose}; on a single frame \cite{RecurrentPose} or a video sequence \cite{3D_Video_Occlusion-Aware}; for a single \cite{CPM} or multiple subjects \cite{OpenPose}.
The main challenges of pose estimation, especially in multi-person settings, are due the large number of degrees of freedom in the human body mechanics and the occurrence of joint occlusions.
To overcome the difficulty of detecting joints under occlusion, it is common for methods to rely on statistical and geometric models to estimate occluded joints \cite{GeometricPose}, \cite{StatisticalPose}.
Anchor poses are also used as a resource to overcome occlusion \cite{LCR-Net}, but this could limit the generalization power of the model and its ability to handle unforeseen poses.

\begin{figure}[t]
\begin{center}
\includegraphics[width=1\linewidth]{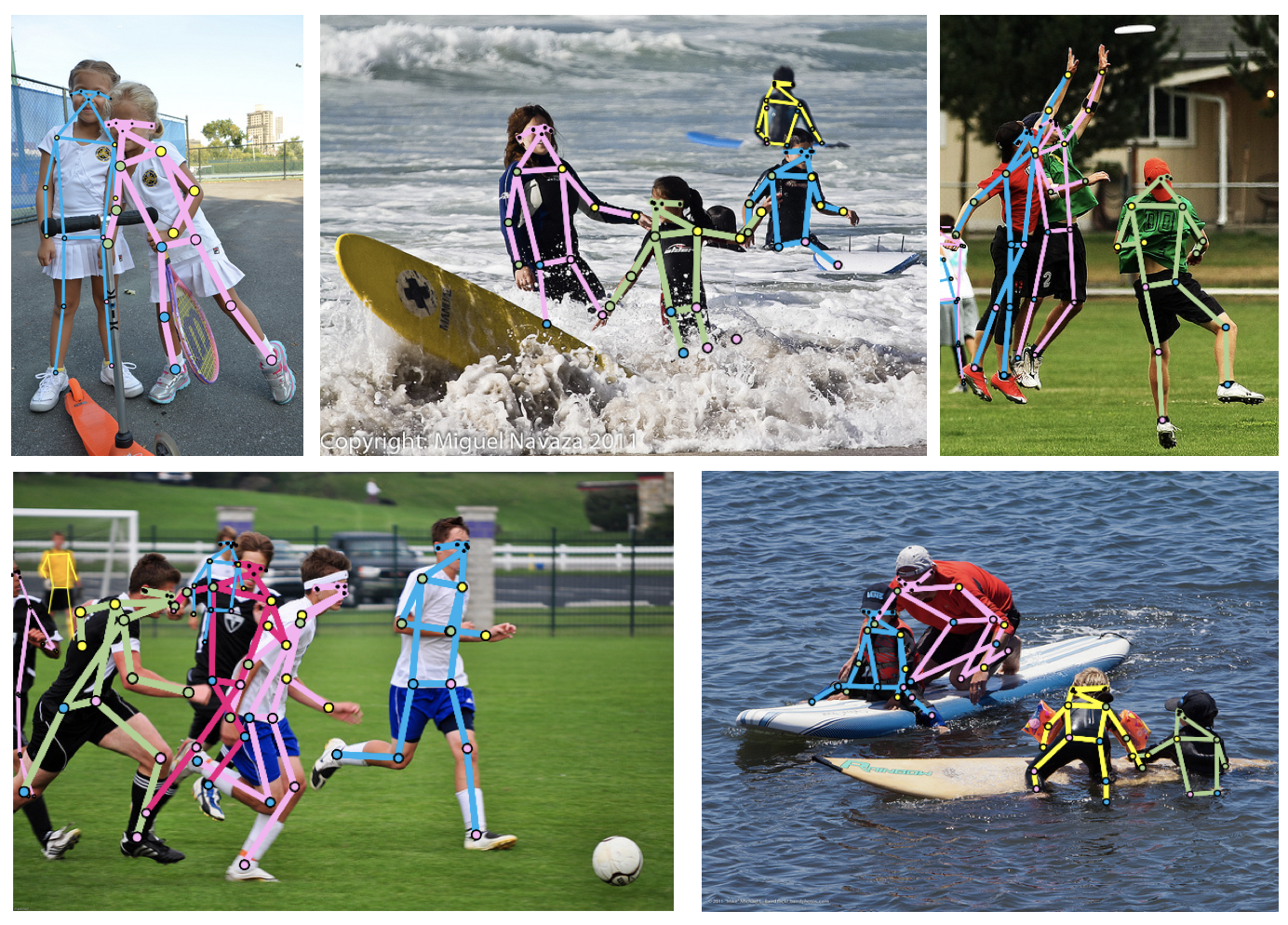}
\end{center}
  \caption{Pose estimation examples with our OmniPose method.}
\label{fig:samples}
\end{figure}

Inspired by recent advances in the use of multi-scale approaches for semantic segmentation \cite{DeepLabv3}, \cite{DilatedConv}, and expanding upon state-of-the-art results
on 2D pose estimation by HRNet \cite{HRNet} and UniPose \cite{UniPose}, we propose OmniPose, an expanded single-stage network that is end-to-end trainable and generates state-of-the-art results without requiring multiple iterations, intermediate supervision, anchor poses or postprocessing.
A main aspect of our novel architecture is an expanded  multi-scale feature representation that combines an improved HRNet feature extractor with advanced Waterfall Atrous Spatial Pooling (WASPv2) module.
Our improved WASPv2 module combines the cascaded approach for atrous convolution with larger Field-of-View (FOV) and is integrated with the network decoder offering significant improvements in accuracy.

Our OmniPose framework predicts the location of multiple people's joints based on contextual information due to the multi-scale feature representation used in our network. The contextual approach allows our network to include the information from the entire frame, and consequently does not require post analysis based on statistical or geometric methods. In addition, the waterfall atrous module, allows a better detection of shapes, resulting in a more accurate estimation of occluded joints.
Examples of pose estimation obtained with our OmniPose method are shown in Figure \ref{fig:samples}.
The main contributions of this paper are the following.

\begin{itemize}
\item We propose the novel OmniPose framework, a single-pass, end-to-end trainable, multi-scale approach that produces state-of-the-art results for multi-person pose estimation.
\vspace{-0.1in}
\item We propose an improved Waterfall module that increases the performance of the network by using a larger field view while maintaining the high resolution of feature maps through the branches of the module. In addition, the WASPv2 module acts simultaneously as feature extractor and decoder, reducing the computational cost and size of the network.
\vspace{-0.1in}
\item The OmniPose framework achieves an increase in performance by incorporating Gaussian heatmap modulation 
that enhances deconvolution operations
in the multi-scale encoder-decoder architecture for a more accurate representation of joint locations and reduction of the quantization error in the network.
\vspace{-0.1in}
\item We propose the novel lightweight OmniPose-Lite architecture that achieves high accuracy results while dramatically decreasing the number of parameters and computational cost of the network by leveraging the size reduction of separable convolutions throughout the network.
\end{itemize}

\section{Related Work}
In recent years,  deep learning methods relying on Convolutional Neural Networks (CNNs) have achieved superior results in human pose estimation \cite{DeepPose}, \cite{CPM}, \cite{OpenPose}, \cite{HRNet}, \cite{LCR-Net} 
over early works \cite{Poselet}, \cite{Articulated}. 
The popular Convolutional Pose Machine (CPM) \cite{CPM} proposed an architecture that refined joint detection via a set of stages in the network.
Building upon \cite{CPM}, Yan et al. integrated the concept of Part Affinity Fields (PAF) resulting in the OpenPose method \cite{OpenPose}. 

Multi-scale representations have been successfully used in backbone structures for pose estimation. Stacked hourglass networks \cite{HourGlass} use cascaded structures of the hourglass method for pose estimation. Expanding on the hourglass structure, the multi-context approach in \cite{Multi-context} relies on an hourglass backbone to perform pose estimation. The original backbone is augmented by
the Hourglass Residual Units (HRU) with the goal of increasing the receptive FOV. Postprocessing with Conditional Random Fields (CRFs) is used to assemble the relations between detected joints. However, the drawback of  CRFs is increased complexity that requires high computational power and reduces speed.

The High-Resolution Network (HRNet) \cite{HRNet} includes both high and low resolution representations.
HRNet benefits from the larger FOV of multi resolution, a capability that we achieve in a simpler fashion with our WASPv2 module. 
An analogous approach to HRNet is used by the Multi-Stage Pose Network (MSPN) \cite{MSPN}, where the HRNet structure is combined with cross-stage feature aggregation and coarse-to-fine supervision.

UniPose \cite{UniPose} combined the bounding box generation and pose estimation in a single one-pass network. This was achieved by the use of WASP module that increases significantly the multi-scale representation and FOV of the network, allowing the method to extract a greater amount of contextual information.

More recently, the HRNet structure was combined with multi-resolution pyramids in \cite{HigherHRNet} to further explore multi-scale features. The Distribution-Aware coordinate Representation of Keypoints (DARK) method \cite{DarkPose} aims to reduce loss during the inference processing of the decoder stage when using an HRNet backbone.

Aiming to use contextual information for pose estimation, the Cascade Prediction Fusion (CPF) \cite{zhang2019human} uses graphical components in order to exploit the context for pose estimation. Similarly, the Cascade Feature Aggregation (CFA) \cite{su2019improvement} uses semantic information for pose with a cascade approach. In a related context, Generative Adversarial Networks (GANs) were used in \cite{SAGAN} to learn dependencies and contextual information for pose.



\begin{figure*}[t]
\begin{center}
\includegraphics[width=1\linewidth]{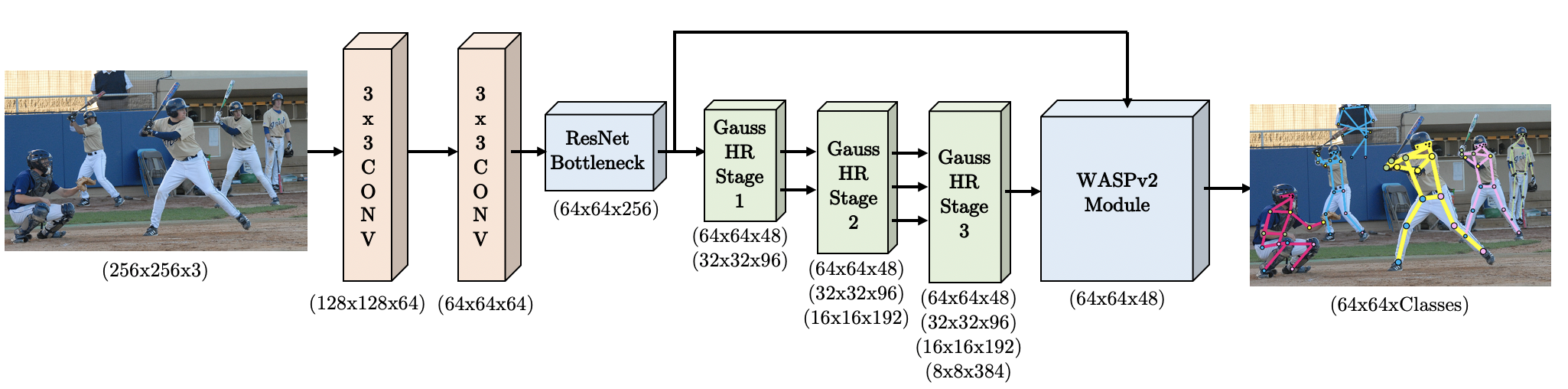}
\end{center}
  \caption{OmniPose framework for multi-person pose estimation. The input color image is fed through the improved HRNet backbone and WASPv2 module to generate one heatmap per joint or class.}
\label{fig:OmniPose}
\end{figure*}

\subsection{Multi-Scale Feature Representations}
A challenge with CNN-based pose estimation, as well as semantic segmentation methods, is a significant reduction of resolution caused by pooling.
Fully Convolutional Networks (FCN) \cite{FCN} addressed
this problem by deploying upsampling strategies across deconvolution layers
that increase the size of feature maps back to the dimensions of the input image. 
In DeepLab, dilated or atrous convolutions \cite{DeepLab} were used to increase the size of the receptive fields in the network and avoid downsampling in a multi-scale framework. 
The Atrous Spacial Pyramid Pooling (ASPP) approach assembles atrous convolutions in four parallel branches with different rates, that are combined by fast bilinear interpolation with an additional factor of eight. This configuration recovers the feature maps in the original image resolution.
The increase in resolution and FOV in the ASPP network can be beneficial for a contextual detection of body parts during pose estimation.

Improving upon \cite{DeepLab}, the waterfall architecture of the WASP module incorporates multi-scale features without immediately parallelizing the input stream \cite{WASP}, \cite{UniPose}. Instead, it creates a waterfall flow by first processing through a filter and then creating a new branch. The WASP module goes beyond the cascade approach by combining the streams from all its branches and average pooling of the original input to achieve a multi-scale representation.

\section{OmniPose Architecture}
The proposed OmniPose framework, illustrated in Figure \ref{fig:OmniPose}, is a single-pass, single output branch network for pose estimation of multiple people instances. OmniPose incorporates improvements in feature representation from multi-scale approaches \cite{HRNet}, \cite{DarkPose} and an encoder-decoder structure combined with spatial pyramid pooling \cite{DeepLabv3+} and our proposed advanced waterfall module (WASPv2).

The processing pipeline of the OmniPose architecture is shown in Figure \ref{fig:OmniPose}. The input image is initially fed into a deep CNN backbone, consisting of our modified version of HRNet \cite{HRNet}. The resultant feature maps are processed by our WASPv2 decoder module that generates K heatmaps, one for each joint, with the corresponding confidence maps. The integrated WASPv2 decoder in our network generates detections for both visible and occluded joints while maintaining the image high resolution through the network.

Our architecture includes several
innovations to increase accuracy. 
The first is the application of atrous convolutions and waterfall architecture of the WASPv2 module, that increases the network's capacity to compute multi-scale contextual information. This is accomplished by the probing of feature maps at multiple rates of dilation during convolutions, resulting in a larger FOV in the encoder.
Our architecture integrates the decoding process within the WASPv2 module without requiring a separate decoder.
Additionally, our network demonstrates good ability to detect shapes by the use of spatial pyramids combined with our modified HRNet feature extraction, as indicated by state-of-the-art (SOTA) results.  Finally, the modularity of the OmniPose framework enables easy implementation and training.


OmniPose leverages the large number of feature maps at multiple scales  in the proposed WASPv2 module. In addition, we improved the results of the backbone network by incorporating gaussian modulated deconvolutions in place of the upsampling operations during transition stages of the original HRNet architecture.
The modified HRNet feature extractor is followed by the improved and integrated multi-scale waterfall configuration of the WASPv2 decoder, which further improves the efficiency of the joint detection with the incorporation of Gaussian heatmap modulation of the decoder stage, and full integration with the decoder module.



\begin{figure}[!ht]
\begin{center}
\includegraphics[width=1\linewidth]{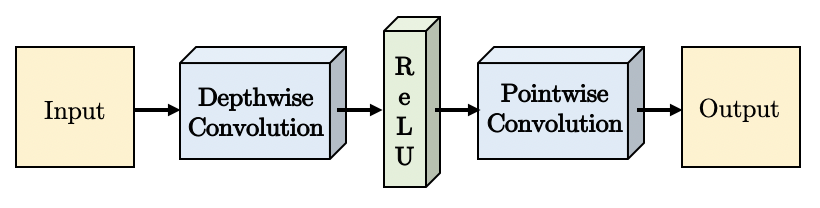}
\end{center}
  \caption{Implementation of our separable convolution. The cascade of depth-wise convolution, ReLU activation, and point-wise convolution replace the standard convolution in order to reduce the number of parameters and computations in the network.}
\label{fig:separable_conv}
\end{figure}

Targeting the reduction of computational cost and number of parameters, we implement separable convolutions replacing the initial two layers of strided convolutions in our model and the atrous convolutions in the WASPv2 module. Figure \ref{fig:separable_conv} demonstrates the implementation of the strided convolution that consists of a spatial (or depth-wise) convolution through the individual channels of the feature maps, followed by a rectified linear unit (ReLU) activation function, and a point-wise convolution to incorporate all the layers of the feature maps.

\begin{figure*}
\begin{center}
\includegraphics[width=1\linewidth]{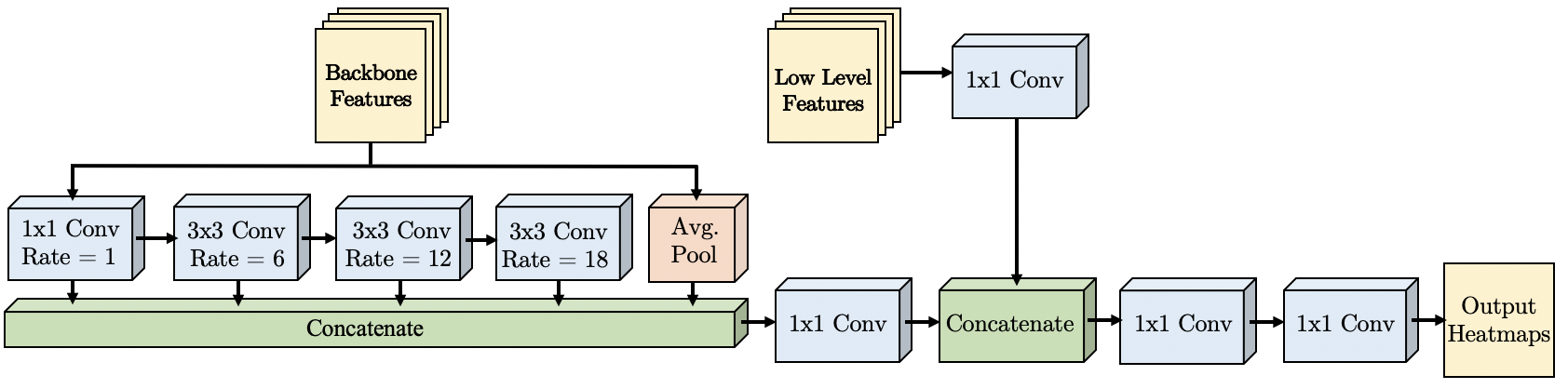}
\end{center}
  \caption{The proposed WASPv2 advanced waterfall module. The inputs are 48 features maps from the modified HRNet backbone and low-level features from the initial layers of the framework.}
\label{fig:WASPv2}
\end{figure*}

\subsection{WASPv2 Module}

The proposed advanced ``Waterfall Atrous Spatial Pyramid'' module, or WASPv2, shown in Figure \ref{fig:WASPv2}, generates an efficient multi-scale representation that helps OmniPose achieve SOTA results. Our improved WASPv2 module expands the feature extraction through its multi-level architecture. It increases the FOV of the network with consistent high resolution processing of the feature maps in all its branches, which contributes to higher accuracy.
In addition, WASPv2 generates the final heatmaps for joint localization without the requirement of an additional decoder module, interpolation or pooling operations.



The WASPv2 architecture relies on atrous convolutions to maintain a large FOV, performing a cascade of atrous convolutions at increasing rates to gain efficiency. In contrast to ASPP \cite{DeepLabv3+}, WASPv2 does not immediately parallelize the input stream. Instead, it creates a waterfall flow by first processing through a filter and then creating a new branch. In addition, WASPv2 goes beyond the cascade approach by combining the streams from all its branches and average pooling of the original input to achieve a multi-scale representation.


Expanding upon the original WASP module \cite{UniPose}, WASPv2 incorporates the decoder in an integrated unit shown in Figure \ref{fig:WASPv2}, and processes both of the waterfall branches with different dilation rates and low-level features in the same higher resolution, resulting in a more accurate and refined response.
The WASPv2 module output $f_{WASPv2}$ is described as follows:
\vspace{-0.1in}
\begin{equation}
    f_{Waterfall} = K_1\circledast(\sum_{i=1}^{4}(K_{d_i}\circledast f_{i-1})+AP(f_{0}))
\end{equation}
\begin{equation}   
    f_{WASPv2} = K_1\circledast(K_1\circledast(K_1\circledast f_{LLF} + f_{Waterfall})
\end{equation}
\noindent 
where $\circledast$ represents convolution, $f_{0}$ is the input feature map, $f_{i}$ is the feature map resulting from the $i^{th}$ atrous convolution, $AP$ is the average pooling operation, $f_{LLF}$ are the low-level feature maps, $K_1$ and $K_{d_i}$ represent convolutions of kernel size $1\times1$ and $3\times3$ with dilations of ${d_i}=[1,6,12,18]$, as shown in Figure \ref{fig:WASPv2}. 
After concatenation, the feature maps are combined with low level features.
The last $1\times1$ convolution brings the number of feature maps down to the final number of joints for the pose estimation.

Differently than the previous version of WASP, our WASPv2 integrates in the same resolution the feature maps from the low-level features and the first part of the waterfall module, converting the score maps from the WASPv2 module to heatmaps corresponding to body joints.
Due to the higher resolution afforded by the modified HRNet backbone, the WASPv2 module directly outputs the final heatmaps without requiring an additional decoder module or need for bilinear interpolations to resize the output to the original input size.

Aiming to reduce the computational complexity and size of the network,
and inspired by \cite{DeepLabv3+}, 
our WASPv2 module implements separable atrous convolutions to its feature extraction waterfall branches. The inclusion of separable atrous convolutions in the WASPv2 module further reduces the number of parameters and computation cost of the framework.

\subsection{Gaussian Heatmap Modulation}
Conventional interpolation or upsampling methods during the decoding stage of the network result in an inevitable loss in resolution and consequently accuracy, limiting the potential of the network.
Motivated by recent results 
with distribution aware modulation \cite{DarkPose}, we include Gaussian heatmap modulation to all interpolation stages of our network,
resulting in a more accurate and robust network that diminishes the
localization error due to interpolation.

Gaussian interpolation allows the network to achieve sub-pixel resolution for peak localization following the anticipated Gaussian pattern of the feature response.
This method results in a smoother response and more accurate peak prediction for joints, by eliminating false positives in noisy responses during the joint detection.

\begin{figure}[th]
\begin{center}
\includegraphics[width=1\linewidth]{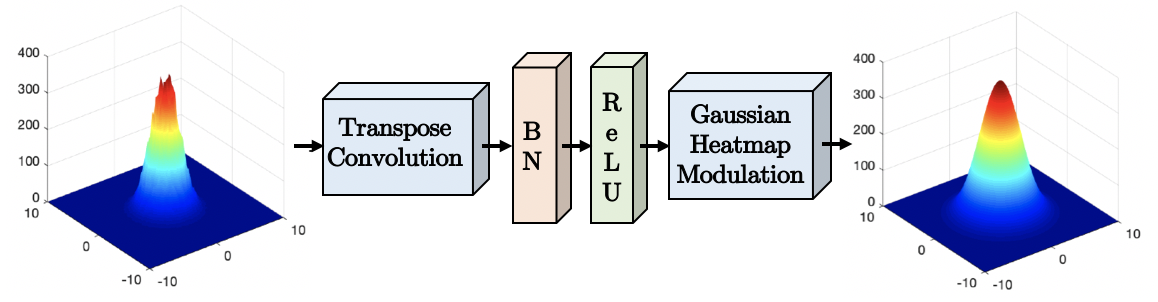}
\end{center}
  \caption{Illustration of the proposed transpose convolution with Gaussian modulation replacing upsampling stages of the network.}
\label{fig:Gauss}
\end{figure}

Figure \ref{fig:Gauss} demonstrates the modularization of a feature map response in our improved HRNet feature extractor.
We utilize a transposed convolution operation of the feature map $f_D$ with a Gaussian kernel $K$, shown in Equation (2), aiming to approximate the response shape to the expected label of the dataset during training.
The feature maps $f_G$ after the Gaussian convolution operation are:
\begin{equation}
    f_G = K \circledast f_D
\end{equation}
\label{convOperation}
\noindent 
This behavior is learned and reproduced by the network during all parts of training, validation, and testing.

Following convolution with the Gaussian kernel, the modulation of the interpolation output is scaled to $f_{G_s}$ by mapping $f_{G}$ to the range of the response of the original feature map $f_{D}$ using:
\begin{equation}
    f_{G_s} = \frac{f_G - min(f_G)}{max(f_G) - min(f_G)}*max(f_D)
\end{equation}
\noindent 

Our Gaussian heatmap modulation approach allows for better localization of the coordinates during the transposed convolutions, by overcoming the quantization error naturally inherited from the increase in resolution. 

\subsection{OmniPose-Lite}
We introduce OmniPose-Lite, a lightweight version of OmniPose that is suitable for mobile and embedded platforms, as it achieves a drastic reduction in memory requirements and operations required for computation. The proposed OmniPose-Lite leverages the reduced computational complexity and size of separable convolutions, inspired by results obtained by MobileNet \cite{MobileNet}.

We implemented separable strided convolutions, as shown in Figure \ref{fig:separable_conv}, for all convolutional layers of the original HRNet backbone, and implemented atrous separable convolutions in the WASPv2 decoder, resulting in a reduction of 74.3\% of the network GFLOPs, from 22.6 GFLOPs to 5.8 GFLOPs required to process an image of size 256x256. In addition, OmniPose-Lite also reduces the number of parameters by 71.4\%, from 67.9M to 19.4M.

The small size of the proposed OmniPose-Lite architecture, in combination with the reduced number of parameters allows the implementation of the OmniPose architecture for mobile applications without a large computational burden.

\section{Datasets}
We performed multi-person experiments on two datasets: Common Objects in Context (COCO) \cite{COCO} and MPII \cite{MPII}.  
The COCO dataset \cite{COCO} is composed of over 200K images containing over 250K instances of the person class. The labelled poses contain 17 keypoints. The dataset is considered a challenging dataset due to the large number of images in a diverse set of scales and occlusion for poses in the wild.

The MPII dataset \cite{MPII} contains approximately 25K images of annotated body joints of over 40K subjects. The images are collected from YouTube videos in 410 everyday activities. The dataset contains frames with joints annotations, head and torso orientations, and body part occlusions.

In order to better train our network for joint detection, ideal Gaussian maps were generated at the joint locations in the ground truth.
These maps are more effective for training than single points at the joint locations, and were used to train our network to generate Gaussian heatmaps corresponding to the location of each joint.
Gaussians with different $\sigma$ values were considered and a value of $\sigma=3$ was adopted, resulting in a well defined Gaussian curve for both the ground truth and predicted outputs. This value of $\sigma$ also allows enough separation between joints in the image.

\section{Experiments}
OmniPose experiments were based on the metrics set by each dataset, and procedures applied by \cite{HRNet}, \cite{HigherHRNet}, and \cite{DarkPose}.

\begin{figure*}[t]
\begin{center}
\includegraphics[width=0.95\linewidth]{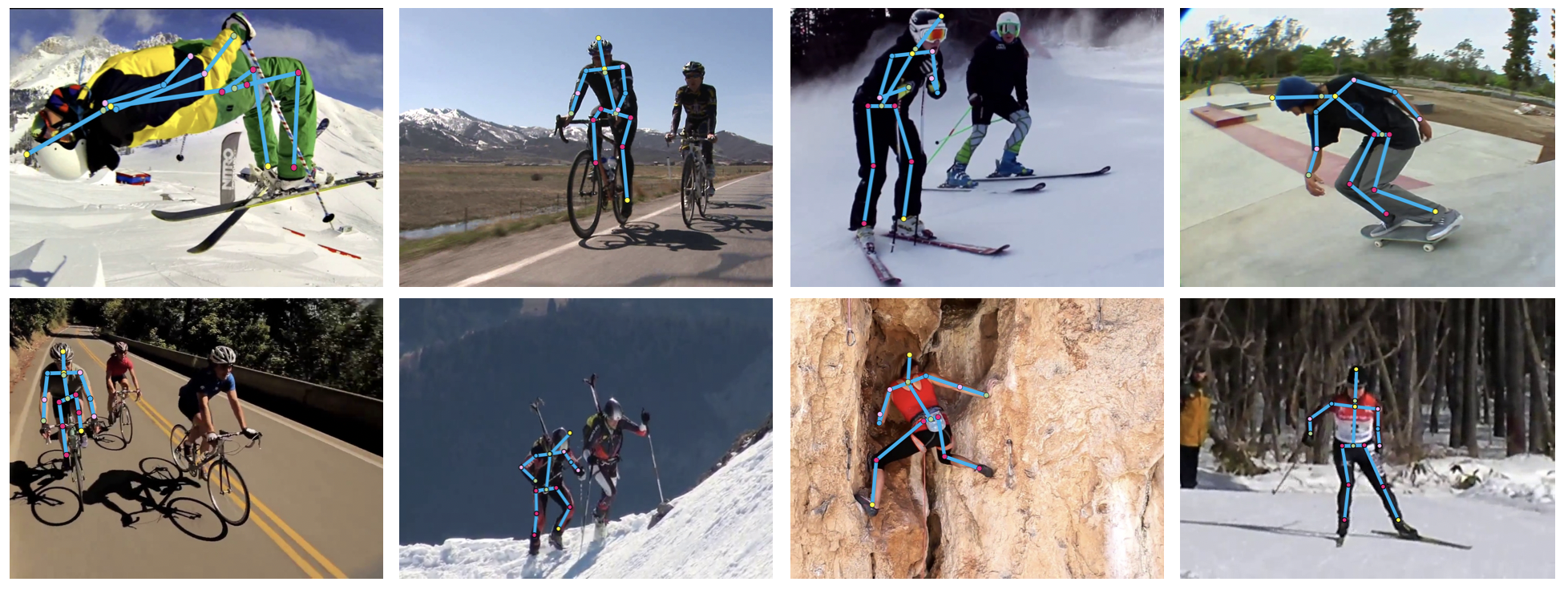}
\end{center}
  \caption{Pose estimation examples using OmniPose with the MPII dataset.}
\label{fig:MPII_sample}
\end{figure*}

\begin{table*}[!ht]
\begin{center}
\begin{tabular}{|c|c|c|c|c|c|c|c|c|c|c|c|}
\hline
\multirow{2}{*}{Method}&Params&\multirow{2}{*}{GFLOPs}&\multirow{2}{*}{Head}&\multirow{2}{*}{Shoulder}&\multirow{2}{*}{Elbow}&\multirow{2}{*}{Wrist}&\multirow{2}{*}{Hip}&\multirow{2}{*}{Knee}&\multirow{2}{*}{Ankle}&PCKh\\
&(M)&&&&&&&&&@0.2\\
\hline\hline
\textbf{OmniPose (WASPv2)}&68.1&22.6&
\textbf{97.4\%}&\textbf{97.1\%}&
\textbf{92.4\%}&\textbf{88.7\%}&
\textbf{91.2\%}&\textbf{89.9\%}&
\textbf{85.8\%}&\textbf{92.3\%}\\
\textbf{OmniPose (WASP)}&68.2&23.0&
97.4\%&96.6\%&
91.9\%&87.2\%&
90.1\%&88.0\%&
83.9\%&91.2\%\\
DarkPose \cite{DarkPose}&63.6&19.5&
97.2\%&95.9\%&
91.2\%&86.7\%&
89.7\%&86.7\%&
84.0\%&90.6\%\\
HRNet \cite{HRNet}&63.6&19.5&
97.1\%&95.9\%&
90.3\%&86.5\%&
89.1\%&87.1\%&
83.3\%&90.3\%\\
\textbf{OmniPose-Lite}&
\textbf{19.4}&
\textbf{5.8}&
96.6\%&95.8\%&
89.1\%&84.3\%&
89.0\%&84.1\%&
79.6\%&89.0\%\\
CMU Pose \cite{OpenPose}&-&-&
92.4\%&90.4\%&
80.9\%&70.8\%&
79.5\%&73.1\%&
66.5\%&79.1\%\\
SPM \cite{SPM}&-&-&
92.0\%&88.5\%&
78.6\%&69.4\%&
77.7\%&73.8\%&
63.9\%&77.7\%\\
RMPE \cite{RMPE}&-&-&
88.4\%&86.5\%&
78.6\%&70.4\%&
74.4\%&73.0\%&
65.8\%&76.7\%\\
\hline
\end{tabular}
\end{center}
\caption{OmniPose results and comparison with SOTA methods for the MPII dataset for validation.}
\label{tab:MPIIval}
\end{table*}

\subsection{Metrics}
For the evaluation of OmniPose, various metrics were used depending on previously reported results and the available ground truth for each dataset.
The first metric used is the Percentage of Correct Keypoints (PCK). This metric considers the prediction of a keypoint correct when a joint detection lies within a certain threshold distance of the ground truth. 
The commonly used threshold of PCKh@0.5 was adopted for the MPII dataset, which refers to a threshold of 50\% of the head diameter.

In the case of the COCO dataset, the evaluation is done based on the Object Keypoint Similarity metric (OKS).

\begin{equation}
    OKS = \frac{(\sum_i{e^{-d_i^2/2s^2k^2_i}})\delta(v_i>0)}{\sum_i\delta(v_i>0)}
\end{equation}

\noindent 
where, $d_i$ is the Euclidian distance between the estimated keypoint and its ground truth, $v_i$ indicates if the keypoint is visible, $s$ is the scale of the corresponding target, and $k_i$ is the falloff control constant.

Since OKS is measured in an analogous form as the intersection over the union (IOU), and following the evaluation framework set by \cite{COCO}, we report OKS as the Average Precision (AP) for the IOUs for all instances between 0.5 and 0.95 ($AP$), at 0.5 ($AP^{50}$) and 0.75 ($AP^{75}$), as well as instances of medium ($AP^M$) and large size ($AP^L$). We also report the Average Recall between 0.5 and 0.95 (AR). 

\subsection{Parameter Selection}
We process the input image in a set of different resolutions, reporting the trade-off of network size and accuracy performance.
For that reason, the batch size varied depending on the size of the dataset images.
We considered different rates of dilation on the WASP module and found that larger rates result in better prediction. A set of dilation rates of $r =$ \{1, 6, 12, 18\} was selected for the WASPv2 module.

We calculate the learning rate based on the step method, where the learning rate started at $10^{-3}$ and was reduced in two steps by an order of magnitude at each steps at 170 and 200 epochs, following procedures set by \cite{DarkPose}. 
All experiments were performed using PyTorch on Ubuntu 16.04. The workstation has an Intel i5-2650 2.20GHz CPU with 16GB of RAM and an NVIDIA Tesla V100 GPU.

\section{Results}
We present OmniPose results on two large datasets and provide comparisons with state-of-the art methods.

\subsection{Experimental results on the MPII dataset}
During our experiments on the MPII dataset, we performed a series of ablation studies to analyze the gains due to different aspects of our method. Table \ref{tab:MPIIanalysis} demonstrates the results for the inclusion of the Gaussian deconvolution modulation (GDM) in the HRNet backbone, and improvements gained by initially using the original WASP module \cite{WASP}, \cite{UniPose}, and then our proposed advanced WASPv2 in combination with the improved HRNet feature extractor.

\begin{table}[!h]
\begin{center}
\begin{tabular}{|c|ccc|c|}
\hline
\multirow{2}{*}{Method}&\multirow{2}{*}{GDM}&\multirow{2}{*}{WASP}&\multirow{2}{*}{WASPv2}&{PCKh}\\
&&&&@0.2\\
\hline\hline
DarkPose \cite{DarkPose}&&&&90.6\%\\
OmniPose&\checkmark&&&91.0\%\\
OmniPose&\checkmark&\checkmark&&91.2\%\\
OmniPose&\checkmark&&\checkmark&92.3\%\\
\hline
\end{tabular}
\end{center}
\caption{Results using different versions of OmniPose and comparison with SOTA for the MPII dataset for validation. GDM represents the use of Gaussian Deconvolution Modulation in the modified HRNet backbone, and WASP and WASPv2 indicates the use of the waterfall modules in the network.}
\label{tab:MPIIanalysis}
\end{table}

\begin{figure*}[ht!]
\begin{center}
\includegraphics[width=1\linewidth]{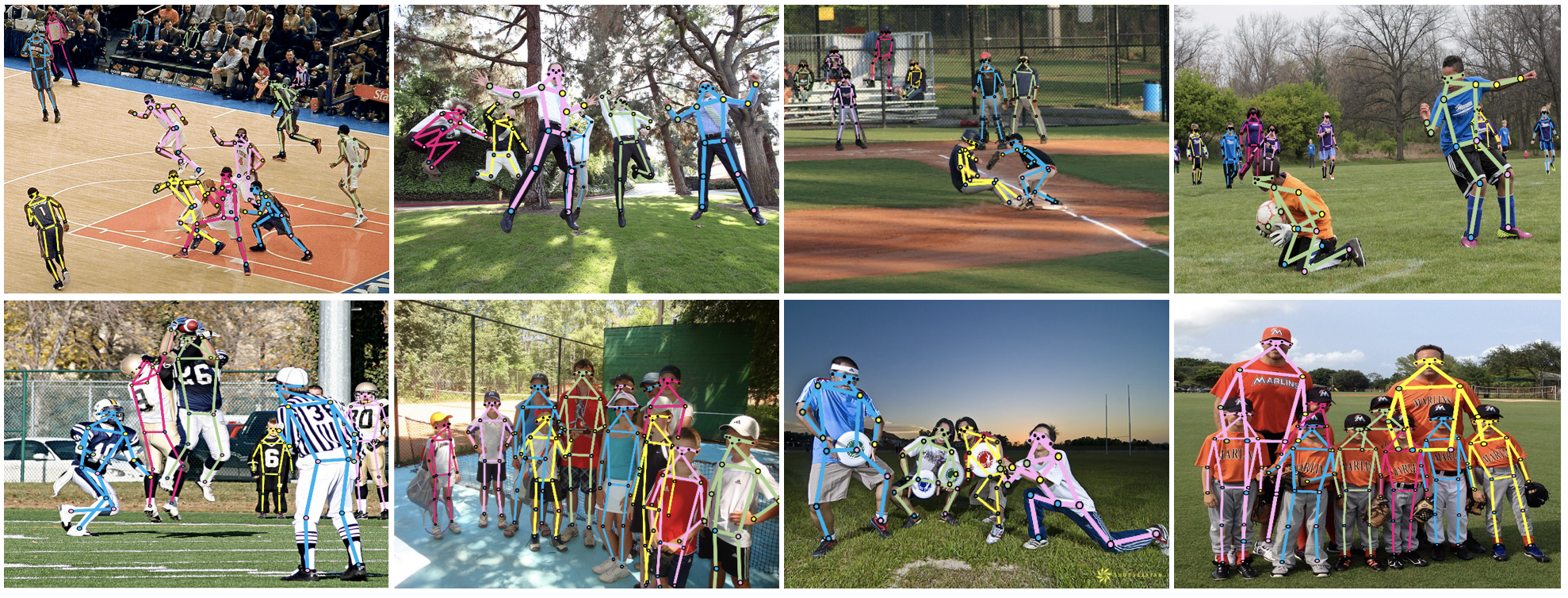}
\end{center}
  \caption{Pose estimation examples using OmniPose with the COCO dataset.}
\label{fig:COCO_sample}
\end{figure*}

\begin{table*}[!ht]
\begin{center}
\begin{tabular}{|c|c|c|c|c|c|c|c|c|c|}
\hline
Method&Input Size&Params (M)&GFLOPs&AP&$AP^{50}$&$AP^{75}$&$AP^{M}$&$AP^{L}$&AR\\
\hline\hline
\textbf{OmniPose (WASPv2)}&384x288&
68.1&37.9&
\textbf{79.5\%}&\textbf{93.6\%}&
\textbf{85.9\%}&\textbf{76.0\%}&
\textbf{84.6\%}&\textbf{81.9}\%\\
OmniPose (WASP)&384x288&
68.2&38.6&
79.2\%&93.6\%&
85.7\%&75.9\%&
84.2\%&81.6\%\\
DarkPose \cite{DarkPose}&384x288&
63.6&32.9&
76.8\%&90.6\%&
83.2\%&72.8\%&
84.0\%&81.7\%\\
HRNet \cite{HRNet}&384x288&
63.6&32.9&
76.3\%&90.8\%
&82.9\%&72.3\%
&83.4\%&81.2\%\\
EvoPose2D \cite{EvoPose2D}&384x288&
7.3&5.6&
75.1\%&90.2\%&
81.9\%&71.5\%&
81.7\%&81.0\%\\
Simple Baseline \cite{SimpleBaseline}&
384x288&68.6&35.6&
74.3\%&89.6\%&
81.1\%&70.5\%&
79.7\%&79.7\%\\
\hline
\end{tabular}
\end{center}
\caption{OmniPose results and comparison with SOTA methods for the COCO dataset for validation.}
\label{tab:COCOval}
\end{table*}

\begin{table*}[!ht]
\begin{center}
\begin{tabular}{|c|c|c|c|c|c|c|c|c|c|}
\hline
Method&Input Size&Params (M)&GFLOPs&AP&$AP^{50}$&$AP^{75}$&$AP^{M}$&$AP^{L}$&AR\\
\hline\hline
\textbf{OmniPose (WASPv2)}&384x288&
68.1&37.9&
\textbf{76.4\%}&92.6\%&
\textbf{83.7\%}&\textbf{72.6\%}&
\textbf{82.6\%}&81.2\%\\
DarkPose \cite{DarkPose}&384x288&
63.6&32.9&
76.2\%&92.5\%&
83.6\%&72.5\%&
82.4\%&81.1\%\\
MSPN \cite{MSPN}&384x288&
120&19.9&
76.1\%&\textbf{93.4\%}&
83.8\%&72.3\%&
81.5\%&\textbf{81.6\%}\\
HRNet \cite{HRNet}&384x288&
63.6&32.9&
75.5\%&92.5\%&
83.3\%&71.9\%&
81.5\%&80.5\%\\
Simple Baseline \cite{SimpleBaseline}&384x288&
68.6&35.6&
73.7\%&91.9\%&
70.3\%&81.1\%&
80.0\%&79.0\%\\
RMPE \cite{RMPE}&320×256&
28.1&26.7&
72.3\%&89.2\%&
79.1\%&68.0\%&
78.6\%&-\\
CPN \cite{CPN}&384x288&
-&-&
72.1\%&91.4\%&
80.0\%&68.7\%&
77.2\%&78.5\%\\
IPR \cite{IPR}&256x256&
45.1&11.0&
67.8\%&88.2\%&
74.8\%&63.9\%&
74.0\%&-\\
G-RMI \cite{G-RMI}&256x256&
42.6&57.0&
64.9\%&85.5\%&
71.3\%&62.3\%&
70.0\%&69.7\\
Mask-RCNN \cite{MaskRCNN}&-&
-&-&
63.1\%&87.3\%&
68.7\%&57.8\%&
71.4\%&-\\
CMU Pose \cite{OpenPose}&-&
-&-&
61.8\%&84.9\%&
57.1\%&67.5\%&
68.2\%&66.5\%\\
\hline
\end{tabular}
\end{center}
\caption{OmniPose results and comparison with SOTA methods for the COCO dataset for test.}
\label{tab:COCOtest}
\end{table*}

Our OmniPose method progressively increases its performance with the addition of innovations, resulting in 1.9\% improvement over DarkPose \cite{DarkPose}. Most significantly, the integration of the enhanced multi-scale extraction with  WASPv2 substantially increases the keypoints detection, particularly for occluded joints.

Following our experiments evaluating the individual contributions of this work, we compared the results of OmniPose with other methods, as shown in Table \ref{tab:MPIIval}.
OmniPose achieved a overall PCKh@0.2 of 92.3\%, showing significant gains in comparison to state-of-the-art. It is significant that OmniPose results in a improvement to previous SOTA methods in all individuals groups of joints for pose estimation, demonstrating the robustness and performance of our framework, particularly to harder to detect joints such as ankles (2.1\% improvement from previous state-of-the-art) and wrists (2.3\% above previous state-of-the-art).
Figure \ref{fig:MPII_sample} demonstrates successful detections on the main person in MPII images. These examples illustrate that OmniPose deals effectively with occlusion, e.g. in the case of the skier.

OmniPose-Lite achieves accuracy of 89.0\% while reducing computational cost by 74.3\% for the MPII validation dataset (Table \ref{tab:MPIIval}).  This  demonstrates its ability to significantly reduce size and computational cost, while maintaining good performance compared to heavier SOTA methods.


\subsection{Experimental results on the COCO dataset}
We next performed training and testing on the COCO dataset, which is more challenging due to the large number of diverse images with multiple people in close proximity, as well as images lacking a person instance.

\begin{figure}[h!]
\begin{center}
\includegraphics[width=1\linewidth]{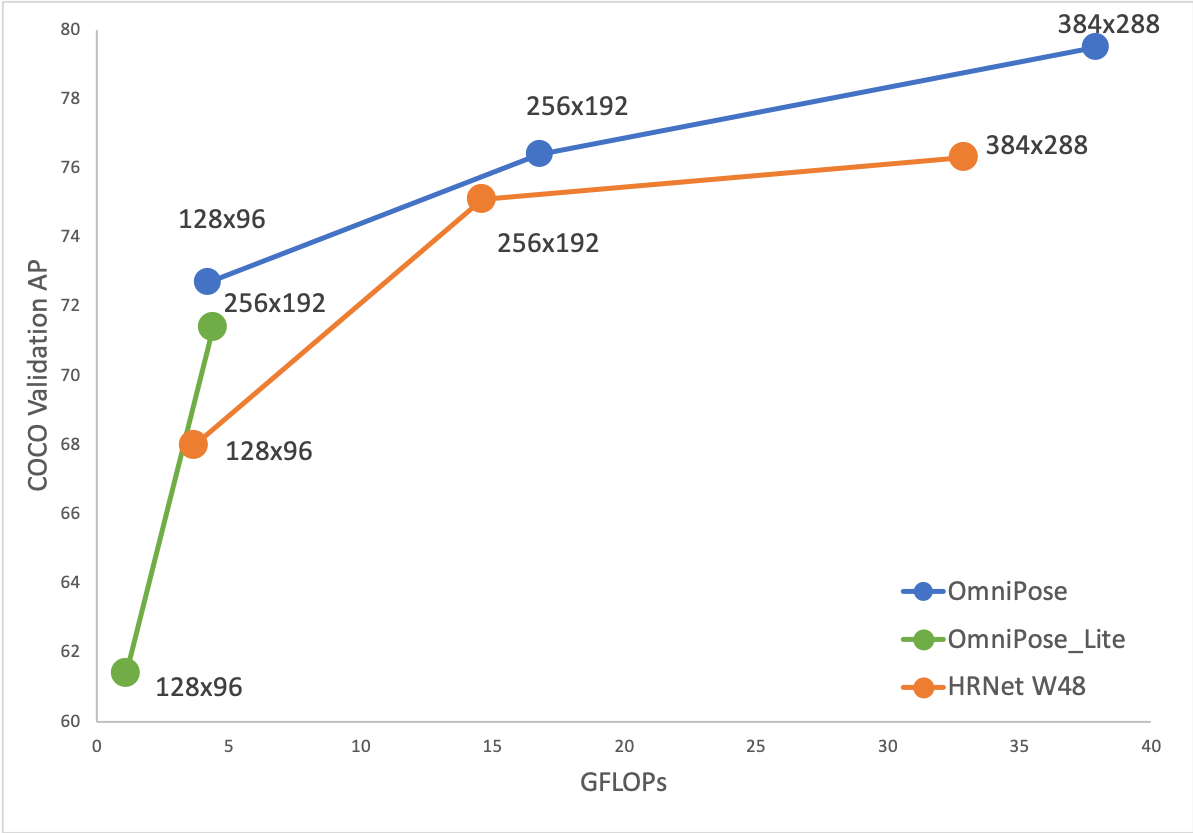}
\end{center}
  \caption{Average Precision comparison of OmniPose to the original HRNet method for different input resolutions.}
\label{fig:APvsGFLOP}
\end{figure}

We performed experiment to compare the proposed improvements of OmniPose with the original HRNet framework.
OmniPose outperforms HRNet in terms of average precision for different input resolutions, as shown in Figure \ref{fig:APvsGFLOP} for 3 different versions of OmniPose: small (128$\times$96), medium (256$\times$192), and large (384$\times$288); as well as lower resolution versions of OmniPose-Lite.
OmniPose demonstrates an increase in performance for all resolutions compared with the original HRNet architecture. The accuracy of the OmniPose framework steadily increases with the increase of the input resolution, but there is a trade-off with processing time due to the larger number of image pixels that are processed in the network.

OmniPose was compared with SOTA methods for the validation set of the COCO dataset. 
The results in Table \ref{tab:COCOval} demonstrate that OmniPose shows significant improvement over the previous SOTA. 
The modification of the HRNet backbone, combined with the WASPv2 module results in an improved accuracy of 79.5\%, a significant increase of 4.2\% compared with the original HRNet, and 7.0\% compared with the baseline model. 

OmniPose improves accuracy for all detection metric sizes and IOU for COCO, as was the case for MPII. Most significantly, in harder detections the AP for person instances of medium size obtained by OmniPose shows an increase of 4.4\% over the previous state-of-the-art.
These results demonstrate the increased capability of OmniPose to estimate harder poses using a reduced number of pixels due to the multi-scale features from the WASPv2 module.

Comparing OmniPose-Lite to lightweight architectures, OmniPose-Lite shows a reduction of size of 12\% while increasing the performance on the COCO validation set by 8.7\% compared to the popular MobileNetV2 \cite{MobileNet}, as shown in Table \ref{tab:COCO_Mobile}. This establishes a significant improvement for lightweight pose estimation methods.

\begin{table}[!ht]
\begin{center}
\begin{tabular}{|c|c|c|c|}
\hline
Method&Input Size&GFLOPs&AP\\
\hline\hline
\textbf{OmniPose-Lite }&256x192&
\textbf{4.4}&\textbf{71.4\%}\\
MobileNetV2 \cite{MobileNet}&256x192&
5.0&65.7\%\\
\hline
\end{tabular}
\end{center}
\caption{Lightweight comparison for the COCO validation dataset.}
\label{tab:COCO_Mobile}
\end{table}

Example results for the validation COCO dataset are shown in Figure \ref{fig:COCO_sample}. 
It is  noticeable  from  these  examples  that  our method identifies the location of symmetric body joints with high precision, providing high accuracy for challenging scenarios, that include multiple instances of people in near proximity and occluded joints, such as ankles and wrists, that are harder to detect.
Challenging conditions include the detection of joints when limbs are not sufficiently separated or occlude each other, where OmniPose demonstrates a robust ability to detect.

We also compared OmniPose with SOTA methods using the COCO test-dev dataset, which contains a significantly larger number of images.
OmniPose achieved a new state-of-the-art performance compared with other methods without the use of additional training data or postprocessing, achieving an average precision of 76.4\%.
Confirming our findings from previous datasets, OmniPose shows the most significant improvements in smaller targets.

\subsection{Single person and video datasets}
We further tested OmniPose on the Leeds Sports Pose (LSP) \cite{LSP} dataset, for single person pose estimation. OmniPose achieved a significant improvement of 5\% from the previous state-of-the-art achieved by UniPose \cite{UniPose}, resulting in a PCK@0.2 of 99.5\% and saturating the pose estimation performance for the LSP dataset. Similarly, running OmniPose on the PennAction dataset for pose estimation in short sports videos \cite{PennAction} shows saturation in performance by achieving state-of-the-art accuracy of 99.4\% PCK.

\section{Conclusion}
We presented the OmniPose framework for multi-person pose estimation.
The OmniPose pipeline utilizes the improved WASPv2 module that features a waterfall flow with a cascade of atrous convolutions and multi-scale representations. 
The OmniPose framework achieves state-of-the-art performance, with an improved HRNet feature extractor utilizing transposed convolutions with Gaussian heatmap modulation, replacing interpolations.
Omnipose is an end-to-end trainable architecture that does not require anchor poses or postprocessing.
The results of the OmniPose framework demonstrated state-of-the-art performance on several datasets using various metrics. 


{\small
\bibliographystyle{ieee_fullname}
\bibliography{egbib}
}

\end{document}